\documentclass{article}

\pdfoutput=1

\usepackage[nonatbib, final]{neurips_2020}

\usepackage[style=numeric, giveninits]{biblatex}

\usepackage[utf8]{inputenc} 
\usepackage[T1]{fontenc}    
\usepackage{hyperref}       
\usepackage{url}            
\usepackage{booktabs}       
\usepackage{amsfonts}       
\usepackage{nicefrac}       
\usepackage{microtype}      
\usepackage{graphicx}
\usepackage{xcolor}

\title{Inspecting state of the art performance and NLP metrics in image-based medical report generation}

\newcommand\authorNumber[1]{\textsuperscript{#1}}

\author{%
  Pablo Pino \authorNumber{1},
  Denis Parra \authorNumber{1},
  Pablo Messina \authorNumber{1},
  Cecilia Besa \authorNumber{2,3},
  Sergio Uribe \authorNumber{2,3}
  \\
  \authorNumber{1} Department of Computer Science, Pontificia Universidad Católica de Chile, \\
  \authorNumber{2} School of Medicine, Pontificia Universidad Católica de Chile, \\
  \authorNumber{3}
  Millennium Nucleus in Cardiovascular Magnetic Resonance, ANID, \\
  \texttt{\{pdpino,pamessina,cbesa,suribe\}@uc.cl},
  \texttt{dparra@ing.puc.cl}

}

\begin{document}

\maketitle

\begin{abstract}
  %
  Several deep learning architectures have been proposed over the last years to deal with the task of generating a written report given an imaging exam as input.
  Most works evaluate the generated reports using standard Natural Language Processing (NLP) metrics (e.g. BLEU, ROUGE),  reporting significant progress. This article contrast this progress by comparing state of the art (SOTA) models against weak baselines. 
  We show that simple and even naive approaches yield near SOTA performance on most traditional NLP metrics.
  We conclude that evaluation methods in this task should be further studied towards correctly measuring clinical accuracy, involving physicians to contribute to this end.
\end{abstract} 

\section{Introduction}

Writing reports from medical images is a everyday labor for radiologists, like the given example of a chest X-ray shown in Figure \ref{figure:iu-xray-example}.
This task has already been addressed with deep learning approaches \cite[e.g.][]{li2019knowledge, liu2019clinically, tian2018ct, wang2018tienet, xue2019features}, typically with encoder-decoder architectures
employing a CNN to extract image features and an RNN for text generation.
In most works, the authors evaluate the model performance using Natural Language Processing (NLP) metrics, such as
BLEU \cite{papineni2002bleu},
ROUGE-L \cite{lin-2004-rouge} and 
CIDEr-D \cite{vedantam2015cider}.
However, it has been discussed that these metrics cannot measure correctness in the medical domain \cite{boagbaselines, liu2019clinically, zhang2019optimizing}, and that a small improvement in them may not implicate an actual quality improvement \cite{mathur-etal-2020-tangled}.
Some works have attempted to design a more appropriate evaluation method, for example, Liu et al. \cite{liu2019clinically} applied the Chexpert Labeler \cite{irvin2019chexpert} over the original and generated reports to classify 14 diseases, and used classification metrics, such as accuracy and ROC-AUC.
In this work, we further inspect some NLP metrics by benchmarking different models against SOTA using the IU X-ray dataset \cite{10.1093/jamia/ocv080}, showing that weak baselines achieve comparable scores. 



\definecolor{otherorange}{rgb}{0.93, 0.57, 0.13}
\definecolor{otherblue}{rgb}{0, 0.5, 1}

\newcommand{\correctTextOne}[1]{\textcolor{otherorange}{#1}}
\newcommand{\wrongTextOne}[1]{\textcolor{otherorange}{#1}}
\newcommand{\correctTextTwo}[1]{\textcolor{otherblue}{#1}}
\newcommand{\wrongTextTwo}[1]{\textcolor{otherblue}{#1}}

\begin{figure}[!ht]
    \centering
    \begin{tabular}{lll}
    \begin{minipage}{.12\textwidth}
        \includegraphics[width=\linewidth]{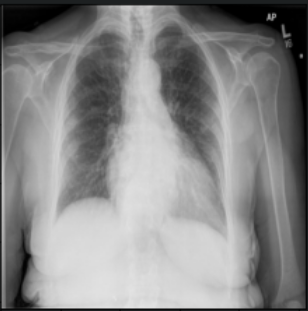}
    \end{minipage}
    &
    \begin{minipage}{.40\textwidth}
        \textbf{Original report (extract):} \\
        The cardiac silhouette \correctTextOne{is enlarged}. The lungs are \correctTextTwo{hyperexpanded with flattening of the bilateral hemidiaphragms}. No pneumothorax or pleural effusion.
    \end{minipage}
    &
    \begin{minipage}{.35\textwidth}
        \textbf{Possible constant report:} \\
        The cardiac silhouette \wrongTextOne{is normal in size and configuration}. The lungs are \wrongTextTwo{clear}. No pneumothorax or pleural effusion.
    \end{minipage}
    \\
    \end{tabular}
    \caption{Example from the IU X-ray dataset \cite{10.1093/jamia/ocv080}: frontal chest X-ray and an extract of the radiologist's report. The third column shows a possible constant report, which obtains high BLEU and ROUGE-L scores of 0.56 and 0.676, but describes incorrect clinical facts in two of its sentences.
    }
    \label{figure:iu-xray-example}
\end{figure}

\section{Methods and Results}

We compared weak baselines and CNN-LSTM architectures against SOTA \cite{liu2019clinically, wang2018tienet, li2019knowledge, xue2019features}.

\textbf{Weak baselines}. Five baselines are implemented.
(1) \textit{Random}: selects a report randomly from the training set.
(2) \textit{Constant}: always outputs the same report, which we built with some of the most common sentences describing no findings (see example in Figure \ref{figure:iu-xray-example}).
(3) \textit{Nearest-neighbor}: looks for the most similar image in the training set and output its report.
(4) \textit{Top-sentences-N} / (5) \textit{Top-words-N}: selects the \textit{N} most common sentences/words of the training set and returns them in random order.

\textbf{CNN-LSTM}. A CNN extracts features from the input image, which are fed to an LSTM that generates the report word by word.
Each input word is passed through an embedding layer.
Two variants are implemented, (1) \textit{CNN-LSTM}: the image features are fed to a fully connected layer to reduce its size and to initialize the LSTM hidden state.
(2) \textit{CNN-LSTM-att}: same as (1), plus using an attention network over the spatial dimension of the image features at each step, and concatenating the attended features with the word embedding input.
The attention network consists of two fully connected layers with a \textit{tanh} activation, following the method used by Tian et al. \cite{tian2018ct}.
We tested several CNNs:
Mobilenet-v2 \cite{howard2017mobilenets},
Resnet-50 \cite{he2016deep},
Densenet-121 \cite{huang2017densely}; the latter performed the best.

\subsection{Results}

Table \ref{table:results} shows the each model's performance in terms of NLP metrics, besides accuracy and ROC-AUC calculated after applying the Chexpert Labeler to the reports (both macro-averaged across 14 diseases).
All metrics range from 0 (worst) to 1 (best), except for CIDEr-D, which reaches a maximum of 10.
The top three results per column are in bold, showing weak baselines achieve scores near to or better than SOTA models, considering BLEU and ROUGE-L.
However, Chexpert metrics show that models are not detecting diseases correctly,
as the baseline AUC for a binary classification task is 0.5.
Notice that because most sentences in the reports describe no abnormalities (about 76\% of the sentences in the dataset), simple baselines can achieve high accuracy.
Also, CIDEr-D shows a strong difference between two SOTA models and naive baselines, though they obtain similar accuracy.
These facts suggest NLP metrics may not be suitable for evaluating the generated reports, and further research should determine how to assess this task automatically.

\newcommand{\firstBest}[1]{\textbf{#1}}
\newcommand{\secondBest}[1]{\textbf{#1}}
\newcommand{\thirdBest}[1]{\textbf{#1}}

\begin{table}[!htbp]
\begin{center}

\caption{\centering Model performance in the IU X-ray dataset \cite{10.1093/jamia/ocv080}.
Results are taken from each paper, except for TieNet which are taken from Liu et al. \cite{liu2019clinically}.
B: BLEU, R-L: ROUGE-L, C-D: CIDEr-D.
}
\begin{tabular}{lccccccccc}
\toprule
Model &  B-1 &  B-2 &  B-3 &  B-4 &  B &  R-L &  C-D &   Acc &  AUC \\
\midrule
Liu et al. \cite{liu2019clinically}  &  0.369 &  0.246 &  0.171 &  0.115 & 0.225 &   \secondBest{0.359} &   \firstBest{1.490} & \secondBest{0.916} &    - \\
TieNet \cite{wang2018tienet} & 0.330 & 0.194 & 0.124 & 0.081 & 0.182 & 0.311 & \secondBest{1.335} & 0.902 & - \\
KERP \cite{li2019knowledge} & \firstBest{0.482} & \secondBest{0.325} & \secondBest{0.226} & \thirdBest{0.162} & \secondBest{0.298} & 0.339 & 0.280 & - & - \\
Xue et al. \cite{xue2019features} & \secondBest{0.477} & \firstBest{0.332} & \firstBest{0.243} & \firstBest{0.189} & \firstBest{0.310} & \firstBest{0.380} & - & - & - \\
\hline 
Constant  &  \thirdBest{0.455} &  \thirdBest{0.312} &  \thirdBest{0.223} &  \secondBest{0.165} & \thirdBest{0.289} &   \thirdBest{0.357} &   \thirdBest{0.293} & \thirdBest{0.915} &    0.500 \\
Random    &  0.362 &  0.197 &  0.117 &  0.075 & 0.188 &   0.264 &   0.112 & 0.894 &    0.508 \\
Nearest-neighbor       &  0.383 &  0.220 &  0.142 &  0.100 & 0.211 &   0.288 &   0.230 & 0.903 &    \firstBest{0.518} \\
Top-sentences-100 &  0.347 &  0.211 &  0.138 &  0.096 & 0.198 &   0.281 &   0.166 & 0.911 &    0.498 \\ 
Top-words-50  &  0.375 &  0.102 &  0.019 &  0.000 & 0.124 &   0.224 &   0.075 & 0.835 &    \secondBest{0.509} \\ 
\hline 
CNN-LSTM      &  0.379 &  0.239 &  0.164 &  0.117 & 0.225 &   0.338 &   0.284 & 0.912 &    0.505 \\
CNN-LSTM-att  &  0.361 &  0.226 &  0.152 &  0.106 & 0.211 &   0.314 &   0.187 & \firstBest{0.918} &    \thirdBest{0.508} \\
\bottomrule
\end{tabular}

    \label{table:results}
    \end{center}
\end{table}

\section{Conclusions and future work}

Given the task of image-based medical report generation, we compared multiple models to the SOTA, using NLP metrics and the Chexpert Labeler.
We found that some weak baselines performed almost as well as some SOTA models in terms of BLEU and ROUGE-L,
suggesting that either these models do not perform so well in a clinical sense, or that these NLP metrics are not able to differentiate automatic methods.
Furthermore, we can raise multiple research questions,
are SOTA models actually performing better than weak baselines in a clinical sense?
Do NLP metrics correlate with medical judgement for this task?
How to directly measure medical correctness in a generated report?
In future work, we will address these questions by asking radiologists to assess generated reports, and we will test the metric MIRQI, which was recently proposed by Zhang et al. \cite{zhang2020radiology} and is designed to measure clinical correctness in the written reports.

\section*{Broader Impact}
This research attempts to generate further discussion on how to measure progress on medical report generation from medical images. Despite the lack of scalable resources to judge generated reports' quality (expert physicians),
we believe the gold standard should consider expert human judgment to assess these methods. Several methods use traditional metrics such as ROUGE or BLEU, but since they have been contested in tasks such as Machine Translation \cite{mathur-etal-2020-tangled}, we suggest to further investigate how appropriate it is to use them for assessing medical report generation. We propose moving forward to metrics actually measuring medical accuracy rather than traditional Natural Language Generation (NLG).
We expect to positively impact this area of research,
by providing new metrics to further impact the progress of the application of NLP to medical reporting.

\begin{ack}
    This work has been supported by the Millennium Institute for Foundational Research on Data (IMFD) and by the Chilean research agency ANID, with the FONDECYT grant 1191791, with the Millennuim Science Initiative Program, NCN17\_129, and by ANID / Scolarship Program / Beca Mag\'ister Nacional / 2020 - 22201476.
\end{ack}


\small
\printbibliography

@article{zhang2020radiology,
  title={When Radiology Report Generation Meets Knowledge Graph},
  author={Zhang, Yixiao and Wang, Xiaosong and Xu, Ziyue and Yu, Qihang and Yuille, Alan and Xu, Daguang},
  journal={arXiv:2002.08277},
  year={2020}
}

@inproceedings{papineni2002bleu,
  title = "{B}leu: a Method for Automatic Evaluation of Machine Translation",
  author={Papineni, Kishore and Roukos, Salim and Ward, Todd and Zhu, Wei-Jing},
  booktitle = "Proc of the 40th Annual Meeting of the ACL",
  pages={311--318},
  year={2002},
  % month = jul,
  % organization={ACL},
  % address = "Philadelphia, Pennsylvania, USA",
%   publisher = "ACL"
}

@inproceedings{lin-2004-rouge,
    title = "{ROUGE}: A Package for Automatic Evaluation of Summaries",
    author = "Lin, Chin-Yew",
    booktitle = "Text Summarization Branches Out",
    % month = jul,
    year = "2004",
    % address = "Barcelona, Spain",
    % publisher = "ACL",
    pages = "74--81",
}

@inproceedings{vedantam2015cider,
  title={Cider: Consensus-based image description evaluation},
  author={Vedantam, Ramakrishna and Lawrence Zitnick, C and Parikh, Devi},
  booktitle = {Proc of the IEEE Conf. on CVPR},
  % month = {06},
  pages={4566--4575},
  year={2015}
}

@inproceedings{irvin2019chexpert,
  title={Chexpert: A large chest radiograph dataset with uncertainty labels and expert comparison},
  author={Irvin, Jeremy and Rajpurkar, Pranav and Ko, Michael and Yu, Yifan and Ciurea-Ilcus, Silviana and Chute, Chris and Marklund, Henrik and Haghgoo, Behzad and Ball, Robyn and Shpanskaya, Katie and others},
  booktitle={Proc of the AAAI Conf. on Artificial Intelligence},
%   volume={33},
  pages={590--597},
  year={2019},
  % month = jul,
%   publisher = {Association for the Advancement of Artificial Intelligence ({AAAI})}
}

@InProceedings{liu2019clinically,
  title = 	 {Clinically Accurate Chest X-Ray Report Generation},
  author =       {Liu, Guanxiong and Hsu, Tzu-Ming Harry and McDermott, Matthew and Boag, Willie and Weng, Wei-Hung and Szolovits, Peter and Ghassemi, Marzyeh},
  pages = 	 {249--269},
  year = 	 {2019},
  booktitle={Machine Learning for Healthcare Conf.},
%   volume = 	 {106},
%   series = 	 {Proc of Machine Learning Research},
  % address = 	 {Ann Arbor, Michigan},
  % month = 	 {08},
  % publisher =    {PMLR}
}

@inproceedings{huang2017densely,
  title={Densely connected convolutional networks},
  author={Huang, Gao and Liu, Zhuang and Van Der Maaten, Laurens and Weinberger, Kilian Q},
  booktitle = {Proc of the IEEE Conf. on CVPR},
  % month = {07},
  pages={4700--4708},
  year={2017}
}

@inproceedings{he2016deep,
  title={Deep residual learning for image recognition},
  author={He, Kaiming and Zhang, Xiangyu and Ren, Shaoqing and Sun, Jian},
  booktitle = {Proc of the IEEE Conf. on CVPR},
  % month = {06},
  pages={770--778},
  year={2016}
}

@article{howard2017mobilenets,
  title={Mobilenets: Efficient convolutional neural networks for mobile vision applications},
  author={Howard, Andrew G and Zhu, Menglong and Chen, Bo and Kalenichenko, Dmitry and Wang, Weijun and Weyand, Tobias and Andreetto, Marco and Adam, Hartwig},
  journal={arXiv:1704.04861},
  year={2017}
}

@InProceedings{boagbaselines,
    title = {{Baselines for Chest X-Ray Report Generation}},
    author = {Boag, William and Hsu, Tzu-Ming Harry and Mcdermott, Matthew and Berner, Gabriela and Alesentzer, Emily and Szolovits, Peter},
    pages = {126--140},
    year = {2020},
    booktitle = {Proc of the Machine Learning for Health NeurIPS Workshop},
    % volume = {116},
    % series = {Proc of Machine Learning Research},
    % address = {},
    % month = {12},
    % publisher = {PMLR}
}

@inproceedings{zhang2019optimizing,
    title = "Optimizing the Factual Correctness of a Summary: A Study of Summarizing Radiology Reports",
    author = "Zhang, Yuhao  and
      Merck, Derek  and
      Tsai, Emily  and
      Manning, Christopher D.  and
      Langlotz, Curtis",
    booktitle = "Proc of the 58th Annual Meeting of the ACL",
    % month = jul,
    year = "2020",
    % address = "Online",
    % publisher = "Association for Computational Linguistics",
    % url = "https://www.aclweb.org/anthology/2020.acl-main.458",
    % doi = "10.18653/v1/2020.acl-main.458",
    pages = "5108--5120",
}

@article{10.1093/jamia/ocv080,
    author = {Demner-Fushman, Dina and Kohli, Marc D. and Rosenman, Marc B. and Shooshan, Sonya E. and Rodriguez, Laritza and Antani, Sameer and Thoma, George R. and McDonald, Clement J.},
    title = "{Preparing a collection of radiology examinations for distribution and retrieval}",
    journal = {Journal of the American Medical Informatics Assoc.},
    % volume = {23},
    % number = {2},
    pages = {304-310},
    year = {2015},
    % month = {07},
    % issn = {1067-5027}
}

@inproceedings{wang2018tienet,
  title={Tienet: Text-image embedding network for common thorax disease classification and reporting in chest x-rays},
  author={Wang, Xiaosong and Peng, Yifan and Lu, Le and Lu, Zhiyong and Summers, Ronald M},
  booktitle = {Proc of the IEEE Conf. on CVPR},
  % month = {06},
  % pages={9049-9058},
  year={2018}
}

@inproceedings{li2019knowledge,
  title={Knowledge-Driven Encode, Retrieve, Paraphrase for Medical Image Report Generation},
  author={Li, Christy Y and Liang, Xiaodan and Hu, Zhiting and Xing, Eric P},
  booktitle={Proc of the AAAI Conf. on Artificial Intelligence},
%   volume={33},
  pages={6666--6673},
  year={2019}
}

@InProceedings{xue2019features,
author="Xue, Yuan
and Huang, Xiaolei",
title="Improved Disease Classification in Chest X-Rays with Transferred Features from Report Generation",
booktitle="IPMI",
year="2019",
% publisher="Springer",
% address="Cham",
pages="125--138",
% isbn="978-3-030-20351-1"
}

@InProceedings{tian2018ct,
author="Tian, Jiang
and Li, Cong
and Shi, Zhongchao
and Xu, Feiyu",
title="A Diagnostic Report Generator from CT Volumes on Liver Tumor with Semi-supervised Attention Mechanism",
booktitle="MICCAI",
year="2018",
% publisher="Springer",
% address="Cham",
pages="702--710",
% isbn="978-3-030-00934-2"
}

@inproceedings{mathur-etal-2020-tangled,
    title = "Tangled up in {BLEU}: Reevaluating the Evaluation of Automatic Machine Translation Evaluation Metrics",
    author = "Mathur, Nitika  and
      Baldwin, Timothy  and
      Cohn, Trevor",
    booktitle = "Proc of the 58th Annual Meeting of the ACL",
    % month = jul,
    year = "2020",
    % address = "Online",
    % publisher = "Association for Computational Linguistics",
    % doi = "10.18653/v1/2020.acl-main.448",
    pages = "4984--4997",
}



\end{document}